\documentclass[letterpaper, 10 pt, conference]{ieeeconf}  

\IEEEoverridecommandlockouts                              

\title{\LARGE \bf
Preparation of Papers for IEEE Sponsored Conferences \& Symposia*
}

\usepackage{macros}

\title{\LARGE \bf \sys: An Efficient MILP Solver for Piecewise-Affine Systems}

\author{
  \authorblockN{Haoze Wu, Min Wu, Dorsa Sadigh, Clark Barrett}
  \authorblockA{Department of Computer Science, Stanford University}
  \authorblockA{\{haozewu, minwu, dorsa, barrett\}@cs.stanford.edu}
}

\begin{document}

\maketitle
\thispagestyle{empty}
\pagestyle{empty}

\begin{abstract}

Piecewise-affine (PWA) systems are widely used for modeling and control of robotics problems including modeling contact dynamics. A common approach is to encode the control problem of the PWA system as a Mixed-Integer Convex Program (MICP), which can be solved by general-purpose off-the-shelf MICP solvers. To mitigate the scalability challenge of solving these MICP problems, existing work focuses on devising efficient and strong formulations of the problems, while less effort has been spent on exploiting their specific structure to develop specialized solvers. The latter is the theme of our work. We focus on efficiently handling one-hot constraints, which are particularly relevant when encoding PWA dynamics. We have implemented our techniques in a tool, \sys, which organically integrates logical reasoning, arithmetic reasoning, and stochastic local search. For a set of PWA control benchmarks, \sys solves more problems, faster, than two state-of-the-art MICP solvers.
\end{abstract}
\section{Background and Motivation}
\label{sec:intro}

\begin{figure}
    \centering
    \includegraphics[height=2.6cm]{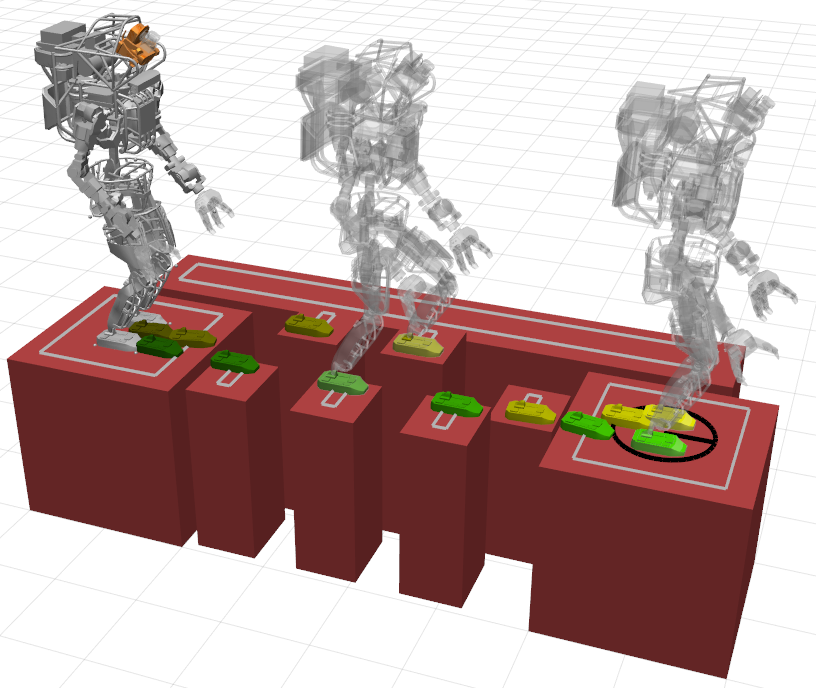}
    \quad
    \includegraphics[height=2.6cm]{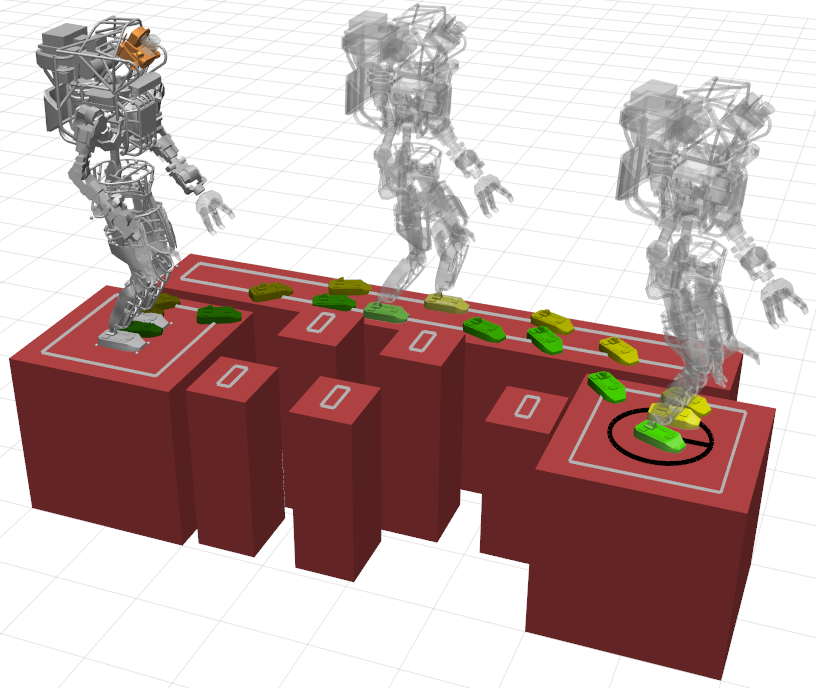}
    \caption{Footstep planning (from~\cite{deits2014footstep}): the robot must find a path using only the available ``stepping stones.'' We evaluate a similar problem later (see Fig.~\ref{fig:stepping_stone}).}
    \label{fig:motivation}
    \vspace{-0.6cm}
\end{figure}

Piecewise-affine (PWA) systems~\cite{sontag1981nonlinear} are widely used to
model highly nonlinear behaviors such as contact dynamics in
robotics. Given an initial condition and a goal, a trajectory that drives the state to the goal while respecting the PWA dynamics and state/control constraints can be obtained by solving a mixed-integer convex programming (MICP) problem.  This approach has seen success in important robotic applications such as push recovery~\cite{han2017feedback} and footstep planning~\cite{deits2014footstep} (as illustrated in Fig.~\ref{fig:motivation}). The MICP approaches are sound and complete for a discrete-time PWA model and a fixed horizon - they find feasible solutions if they exist. However, completeness comes at a high computational cost due to the inherent complexity of solving MICP problems.

Previous work on controlling PWA systems has focused on obtaining efficient (i.e., fewer variables/constraints) and strong (i.e., tighter convex relaxation) formulations of the MICP problems in different application scenarios~\cite{andrikopoulos2013piecewise,marcucci2019mixed,han2017feedback,marcucci2021shortest,deits2014footstep,aceituno2017simultaneous,landry2016aggressive}. The actual solving is typically off-loaded to \emph{general-purpose} off-the-shelf MICP solvers~\cite{mosek,cplex,gurobi,scip}. While the performance of off-the-shelf solvers has improved dramatically in the past decade, these solvers were originally developed with non-robotic applications (e.g., operations research) in mind. Little has been done to tailor solvers specifically to robotics applications. A natural question is then: \textbf{can we exploit the structure of problems arising from PWA systems to design specialized MICP solvers that are faster than general-purpose ones?}

In MICP encodings of problems from the planning domain, logical constraints are typically encoded using arithmetic. When it comes to PWA dynamics, a ubiquitous logical constraint type is the \emph{one-hot constraint}, which encodes the fact that at any time ``the system is in exactly one mode.'' For example, each footstep of the robot in Fig.~\ref{fig:motivation} must be on exactly one of the ``stepping stones.''

Another way to encode one-hot constraints, however, is to use propositional logic directly.
\emph{Our key insight is that reasoning about modes explicitly at the logical level can be beneficial.} To get some intuition for why this is the case, suppose we know that a certain mode combination is infeasible.  To rule out this mode combination, we could either encode it using integer arithmetic constraints or as a single disjunction at the propositional logic level. Empirically, it can be observed that the addition of a few hundred arithmetic (linear) constraints can significantly increase the runtime of a solver.  At the same time, Boolean satisfiability (SAT) solvers can process thousands of disjunctions in milliseconds.

\begin{figure*}[t!]
    \centering
    \includegraphics[width=0.85\textwidth]{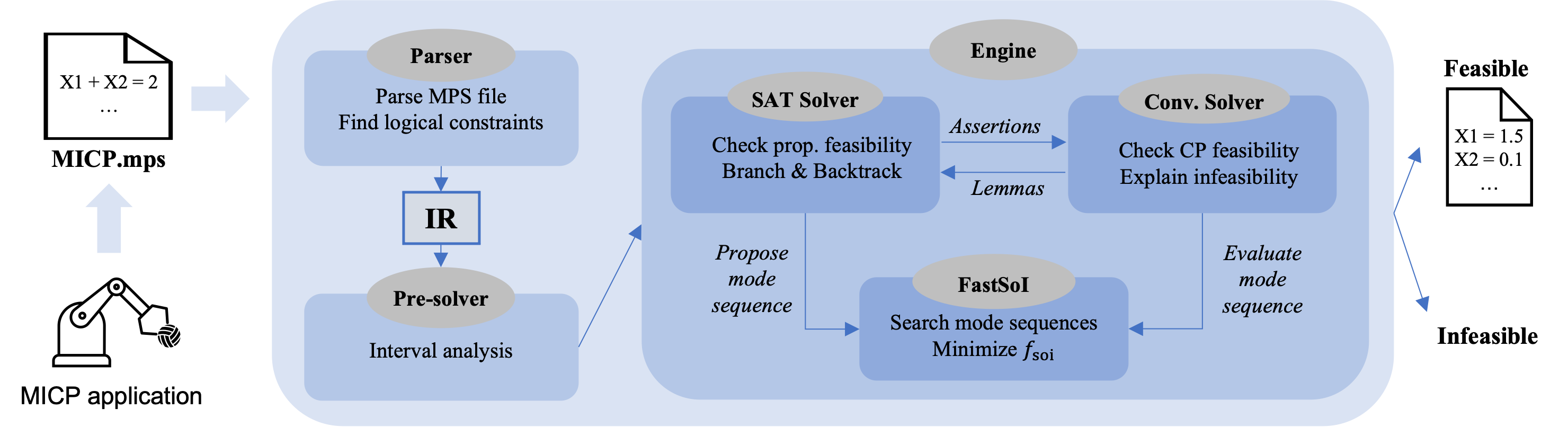}
    \vspace{-1mm}
    \caption{Architectural overview of \sys.}
    \label{fig:overview}
\vspace{-0.6cm}
\end{figure*}

A tight integration of propositional and theory (e.g., arithmetic) reasoning is at the heart of the highly successful satisfiability modulo theories (SMT) paradigm~\cite{barrett2018satisfiability}, with most implementations based on the popular DPLL(T) framework~\cite{ganzinger2004dpll}. In this paper, we adapt the DPLL(T) framework for the setting of PWA planning. In particular, we develop a sound and complete DPLL(T) procedure which integrates propositional reasoning with reasoning about \emph{mixed-integer linear programming} (MILP) problems (a subset of MICP problems) with one-hot constraints.

The underlying convex solver in our approach can only operate on convex relaxations of the one-hot constraints, i.e., integer variables are relaxed to real variables.
To further tailor our solver to our problem domain, 
we propose to ``softly'' guide the convex solver with information about the precise one-hot constraints. Inspired by the sum-of-infeasibilities method in convex optimization~\cite{boyd2004convex}, we define a cost function which represents the degree to which the current solution violates the one-hot constraints. If an assignment is found with cost zero, then not only is the assignment a solution for the convex relaxation, but it also solves the precise problem.

The aforementioned cost function is concave piecewise-linear, which is challenging to minimize directly. We observe, however, that for any specific mode sequence, the system collapses into a set of \emph{linear} constraints, which can be optimized by an LP solver. Minimizing the linear cost function provides a way to evaluate ``how feasible'' the corresponding mode sequence is.  Leveraging this insight, we propose to use Markov chain Monte Carlo (MCMC) sampling to efficiently navigate towards mode sequences at the global minimum of the cost function. In addition, we propose a novel propagation-based proposal strategy for MCMC sampling, which guarantees that the sampled mode sequence is 1) non-repetitive; and 2) does not match any known infeasible mode combinations. 

Our end result is a specialized solver, \sys,\footnote{\sys is available at \href{https://github.com/stanford-centaur/Soy}{https://github.com/stanford-centaur/Soy}} that combines the strength of SMT, MILP, and stochastic local search to efficiently reason about PWA systems. An overview of \sys is shown in Fig.~\ref{fig:overview} and explained in detail in Sec.~\ref{sec:overview}.
While \sys is still an early prototype, it can already be used to solve PWA control problems appearing in the literature faster than was previously possible using existing MILP solvers alone. The closest related work is \cite{shoukry2017smc}, which also shows that combining logical and arithmetic reasoning can be beneficial. Our work goes further by proposing domain-specific solutions for PWA dynamics, combining complete search with local search, and implementing a tool that is friendly to practitioners accustomed to using MICP solvers.

To summarize, our contributions include:
\begin{enumerate}
    \item an instantiation of the DPLL(T) framework for MILP problems with one-hot constraints;
    \item \deepsoi, a novel local search procedure based on the sum-of-infeasibilities method and MCMC sampling;
    \item a propagation-based strategy for MCMC sampling;  
    \item \sys, a specialized MILP solver for PWA control that combines the proposed techniques;
    \item an evaluation of \sys on PWA-control benchmarks.
\end{enumerate}


\section{Preliminaries and Definitions}
\label{sec:prelim}

\smallskip\noindent \textbf{Feasibility of MILP.} Let $\realvars$ be a set of \emph{real variables}. A \emph{linear constraint} has the form $\sum_{x_i\in \realvars}c_i \cdot x_i \bowtie d$, where $\bowtie\ \in \{\leq, <, =\}$ and the $c_i$'s and $d$ are rational constants. An \emph{integral constraint} has the form $x \in \Z$, and a \emph{binary constraint} has the form $x\in\{0, 1\}$, where $x\in \realvars$.
A \emph{solution} $\solution:\realvars\mapsto\R$ is a mapping from variables to real values. $\alpha$ is a \emph{feasible} solution for a set of constraints $\phi$, written $\alpha\models \phi$, if replacing each variable $x$ in $\phi$ by $\alpha(x)$ results in a set of true statements. If no feasible solution exists for $\phi$, $\phi$ is \emph{infeasible}. The MILP feasibility problem is to determine whether a feasible solution exists for a set of constraints.

A \emph{convex relaxation} $\rlx{\phi}$ of $\phi$ can be obtained by dropping all the integral constraints in $\phi$ and replacing binary constraints $x\in\{0,1\}$ with $0\le x\le 1$. The feasibility of $\rlx{\phi}$ can be determined with convex optimization (e.g., the simplex algorithm~\cite{dantzig1955generalized}). If $\rlx{\phi}$ is infeasible, then $\phi$ is infeasible. But the reverse is not true, because a feasible solution to $\rlx{\phi}$ might not satisfy the integral or binary constraints.

\smallskip\noindent \textbf{One-hot constraints.} One-hot constraints can encode requirements like ``the system must be in exactly one mode.'' For example, in locomotion, the modes could correspond to the regions where the robot can step~\cite{deits2014footstep}; and in manipulation, the modes could correspond to disjoint contact scenarios (e.g., contact and no contact)~\cite{marcucci2019mixed}. A one-hot constraint can be encoded over a set $\binvars$ of real variables as follows.

\begin{align}\label{eq:onehotA}
\onehot(\binvars) := \Big(\sum_{x_i \in \binvars} x_i = 1\Big) \land \Big(\bigwedge_{x_i \in \binvars} x_i \in \{0, 1\}\Big)    
\end{align}
Equation~\ref{eq:onehotA} states that exactly one variable in $\binvars$ is 1 and the rest must be 0. For a PWA system with $m$ modes, $\binvars$ would contain $m$ variables, each corresponding to one mode. Alternatively, the one-hot constraint can be encoded in propositional logic. One way to do this is by introducing a set $\pvars=\{p_1, \ldots, p_m\}$ of propositional variables, one for each mode.  The constraint is then:
\begin{align}\label{eq:onehotL}
\onehot_L(\pvars) := \big(\bigvee_{p\in \pvars} p\big) \land \big(\bigwedge_{1\leq i < j \leq m}(\neg p_i \lor \neg p_j)\big)
\end{align}
The first conjunct requires at least one variable in $\pvars$ to be \true, and the rest enforce that at most one variable in $\pvars$ is \true. In order to be able to use both arithmetic and logical constraints together, we need a way to connect the variables.  We assume a bijection $\prop:\binvars\mapsto\pvars$ from real variables to the corresponding propositional variables (e.g., $\prop(x_i) = p_i$).

\smallskip \noindent \textbf{PWA Control as MILP.} When each mode is a polytope (i.e., a bounded polyhedron), PWA dynamics can be encoded with one-hot constraints and linear constraints.
A PWA control problem is defined with respect to an initial condition, a goal, and a horizon $T$.  The question is whether the system can reach the goal within $T$ steps starting from the initial condition.  This can be encoded with a constraint of the form: 
\begin{equation}\label{eq:pwamilp}
\phi := \C_{init}(\realvars^1) \land \bigwedge_{t=1}^{T}(\C_{pwa}(\realvars^t) \land \onehot(\binvars^t))  \land \C_{goal}(\realvars^T)
\end{equation}
where $\realvars^t \in \realvars$ are real variables for step $t$ and $\binvars^t\subset\realvars^t$ correspond to the modes at step $t$.
$\C_{init}$ describes the initial state, and $\C_{goal}$ describes the goal. In this paper, we assume $\C_{init}$ and $\C_{goal}$ only contain linear constraints, so $\phi$ is a set of linear and one-hot constraints. Solving the PWA control problem amounts to checking the feasibility of $\phi$

\smallskip \noindent \textbf{Sum-of-Infeasibilities.} In convex optimization~\cite{boyd2004convex,king2013simplex},
the \emph{sum-of-infeasibilities} (SoI) method can be used to check the feasibility of a set of linear constraints: the feasibility problem is cast as an optimization problem, with a cost function $f(\realvars)$ representing the total \emph{violation} of the constraints by the current solution (e.g.., the sum of the distances from each out-of-bounds variable to its closest bound). The lower bound of $f$ is 0 and is achieved only if the current solution is feasible. In Section~\ref{sec:soi}, below, we build on this idea, proposing a cost function $\soi(\realvars)$ that represents the total violation of the one-hot constraints by the current solution and a stochastic minimization solution.

\section{Checking Feasibility}
\label{sec:dpllt}

\begin{algorithm}[t]
  \small
  \begin{algorithmic}[1]
    \State {\bfseries Input:} linear constraints $\C$, one hot constraints $\O$
    \State {\bfseries Output:} $\feas/\infeas$
    \Function{CheckFeas}{$\C, \O$}
    \State {$\E \mapsto \{\onehot_L(\prop(\binvars))\ |\ \onehot(\binvars)\in \O\}$} \label{line:initE}
    \State {$r, \cdot \mapsto \recdpllt(\C, \O, \emptyset, \E)$} \label{line:recCall}
    \State \textbf{return} r
    \EndFunction
    \State
    \State {\bfseries Input:} $\C$, $\O$, decisions $\D$, and propositional constraints $\E$
    \State {\bfseries Output:} $\feas/\infeas$ and theory lemmas $\L$
    \Function{RecCheckFeas}{$\C, \O, \D, \E$}
    \If {$\neg \satProc(\E)$} \textbf{return} $\infeas, \varnothing$ \label{line:checksat}
    \EndIf
    \State {$r, \alpha, \L \mapsto \convProc(\C \cup \rlx{\O} \cup \D)$} \label{line:checkconv}
    \State {$\L \mapsto \L \cap \D$}
    \If {$r = \infeas$} \textbf{return} $\infeas, \L$  
    \ElsIf {$\alpha \models (\C \cup \O)$} \textbf{return} $\feas, \varnothing$ \label{line:satfound}
    \EndIf
    \For {$\tup{\O_i, \D_i, \E_i} \in \branch(\O, \E)$} \label{line:branch}
    \State {$r_i, \L_i \mapsto \recdpllt(\C, \O_i, \D_i, \E_i \cup \neg\prop(\L))$} \label{line:solvesub}
    \If {$r_i = \feas$} \textbf{return} $\feas, \varnothing$  \label{line:satfoundrec}
    \EndIf
    \State {$\L \mapsto \L \cup \L_i$} \label{line:augmentlemma}
    \EndFor 
    \State \textbf{return} $\infeas, \L$ \label{line:unsatfound}
    \EndFunction
  \end{algorithmic}
  \caption{Complete feasibility search.\label{alg:complete}}
\end{algorithm}

DPLL(T) is a framework for solving SMT problems.\footnote{Here, we touch upon DPLL(T) at a high level;
a detailed presentation can be found in \cite{barrett2018satisfiability}.}  A DPLL(T)-like procedure for MILP problems with one-hot constraints is shown in Alg.~\ref{alg:complete}. 
%
The procedure takes as inputs the linear constraints $\C$ and the one-hot constraints $\O$ and checks if $\C\cup\O$ is feasible. During the solving process, it accumulates new information about the one-hot constraints at the propositional level, which it stores as a set of propositional constraints, \E, initialized with a propositional encoding of the one-hot constraints as described in \eqref{eq:onehotL} (Line~\ref{line:initE}). The following invariant is preserved throughout the execution:
\begin{condition}
If $\C\cup \O$ is feasible, then $\E$ is satisfiable.\footnote{A propositional formula is satisfiable if there exists an assignment to its variables that makes the formula true.}
\label{cond:consistentE}
\end{condition}

$\dpllt$ invokes the recursive function \recdpllt with input arguments $\C$ and $\O$, an empty set of decisions, and $\E$. \recdpllt first checks the satisfiability of $\E$ (Line~\ref{line:checksat}). If it is unsatisfiable, then due to Condition~\ref{cond:consistentE}, $\C \cup\O$ must be infeasible. If $\E$ is satisfiable, we check the feasibility of the convex relaxation $\C \cup \rlx{O} \cup \D$ with $\convProc$ (Line~\ref{line:checkconv}).

$\convProc$ calls a convex feasibility checker with the capability of generating \emph{explanations} in the case of infeasibility. An explanation is an (ideally minimal) infeasible subset of the input constraints~\cite{barrett2018satisfiability}.  The first output of the method is either \feas or \infeas, indicating whether the input is feasible. If so, a feasible solution $\alpha$ is returned and $\L$ is empty. If not, $\L$ contains an explanation.  We only care about the part of the explanation coming from the decisions in $\D$, so we restrict $\L$ accordingly.

For example, suppose $\D=\{x_i=1, x_j=1, x_k=1\}$, corresponding to a certain combination of modes. In addition to deducing that this mode combination is infeasible, the convex procedure might further deduce that $\C \cup \rlx{\O} \cup \{x_i = 1, x_k = 1\}$ is already infeasible. In this case, we would get $\L = \{x_i = 1, x_k = 1\}$. Efficiently generating explanations for infeasible linear constraints is a well-studied problem (see~\cite{gleeson1990identifying}).  Explanations, also called \emph{theory lemmas}, can be used to prune the search space.

The pruning could be done at the arithmetic level by adding a linear constraint $x_i  + x_k < 2$. However, as we accumulate lemmas during the search, this could lead to a drastic slowdown of the convex procedure. An alternative approach (and the one we take) is to record this information as a propositional constraint $\neg\prop(l):= \neg (\prop(x_i)\land \prop(x_k))$ and rely on $\satProc$ (which in general is much faster than the convex solver) to rule out infeasible mode combinations.

If \convProc finds a feasible solution $\alpha$ that is also a feasible solution to the precise constraints $\C \cup \O$, then the solver returns \feas (Line~\ref{line:satfound}). If $\alpha$ does not satisfy the precise constraints, the analysis is inconclusive, and branching is required to make progress (Line \ref{line:branch}): the \branch method selects one of the one-hot constraints in $\O$ and performs case analysis on it. For example, suppose it chooses $\onehot(\{x_1, x_2\})$. Then, \branch will return: 
\begin{align*}
\{\tup{\O \backslash \{\onehot\}, \{x_1 = 1, x_2=0\}, \E \cup \{\prop(x_1) \land \neg\prop(x_2)}, \\
\tup{\O \backslash \{\onehot\}, \{x_1 = 0, x_2=1\}, \E \cup \{\neg\prop(x_1) \land \prop(x_2)}\}.  
\end{align*}

The procedure iteratively solves each of the sub-problems (Lines~\ref{line:branch}-\ref{line:augmentlemma}), accumulating theory lemmas (Line~\ref{line:augmentlemma}), and using them for later iterations (Line~\ref{line:solvesub}). 
The procedure returns \feas if one of the sub-problems returns \feas. Otherwise, the procedure returns \infeas along with all the detected theory lemmas.


\section{Sum of Infeasibilities for Mode Sequences}
\label{sec:soi}

Alg.~\ref{alg:complete} is sound and complete. It returns \feas if and only if there is a feasible solution satisfying both the linear constraints and the precise one-hot constraints. However, since \convProc operates on convex relaxations and is unaware of the binary constraints, it can find spurious solutions that violate the binary constraints, leading to costly branching.
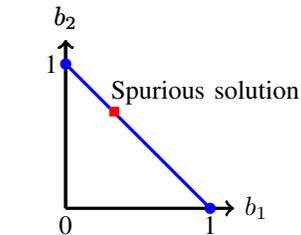
\begin{wrapfigure}{r}{0.25\textwidth}
\vspace{-5mm}
\centering
\begin{tikzpicture}[scale=0.64]
\draw[->, very thick] (0, 0) -- (3.5, 0) node[right] {$b_1$};
\draw[->, very thick] (0, 0) -- (0, 3.5) node[above] {$b_2$};
\draw[->, very thick] (0, 0) -- (0, 3.5) node[above] {$b_2$};
\draw[domain=0:3, very thick, variable=\x, blue] plot ({\x}, {3-\x});


\node at (0, -0.35) {0};
\node at (3, -0.35) {1};
\node at (-0.3, 3) {1};
\draw[draw=red, fill=red] (0.92,1.92) rectangle ++(0.18,0.18);
\node at (2.9, 2.4) {Spurious solution};
\filldraw[blue] (0, 3) circle (3pt);
\filldraw[blue] (3, 0) circle (3pt);
\end{tikzpicture}
\vspace{-3mm}
\caption{Solutions found by convex solver can be spurious.}
\label{fig:onehot}
\vspace{-2mm}
\end{wrapfigure}

Fig.~\ref{fig:onehot} illustrates this phenomenon for a one-hot constraint with two modes (represented by $b_1$ and $b_2$). In its convex relaxation, $(b_1, b_2)$ can lie anywhere on the line segment in the figure.

Ideally, we want the convex optimization procedure to be more aware of the precise one-hot constraints so that it can avoid branching. We propose to achieve this by ``softly'' guiding the convex procedure with a cost function \soi, called the sum-of-infeasibilities, that represents the violation of the one-hot constraints. Minimizing \soi over the convex relaxation naturally leads to a feasible solution that satisfies both the convex relaxation and the precise integral requirements. We next introduce our \soi function for one-hot constraints, consider the challenge of its minimization, and present a stochastic local search solution.

\subsection{The Sum of Infeasibilities}
As mentioned above, in convex optimization, a sum-of-infeasibilities function represents how much the current solution violates the convex constraints. Here, we build on this idea by introducing a cost function $\soi$, which computes the sum of errors introduced by the convex relaxation of one-hot constraints.\footnote{Similar ideas have been used to reason about different types of non-linear constraints in a different setting~\cite{wu2022efficient}.}
$\soi$ needs to meet the following condition:
\begin{condition}
  \label{cond:soi}
  Given a set of linear constraints $\C$ and a set of one-hot constraints $\O$ defined over variables $\realvars$, a solution $\alpha$ is feasible for $\phi:= \C\cup \O$ iff
  $\alpha$ is a feasible solution to $\rlx{\phi}$ and $\soi(\alpha) \leq 0$. \label{condition:soi}
\end{condition}
If Condition~\ref{cond:soi} is met, then feasibility of $\phi$ reduces to
the following minimization problem:
\begin{equation}
  \begin{aligned}
    \minimize_{\realvars} \quad &\soi(\realvars)\\
    \textrm{subject to} \quad &\rlx{\phi}
  \end{aligned}
  \label{eq:min}
\end{equation}

To formulate \soi for a problem with one-hot constraints, we first define the error in a single one-hot constraint $\onehot(\binvars)$ as:
$\vio(\binvars) = 1 - \max(\binvars)$.
Note that $\vio(\binvars)$ subject to $\rlx{o}$ is non-negative and $\vio(\binvars) = 0$ iff $o$ is satisfied.
We  define $\soi$ as the sum of errors in each individual one-hot constraint:
\begin{equation}
  \soi = \sum_{\onehot(\binvars) \in \O}\vio(\binvars)
  \label{eq:soi}
\end{equation}

\begin{theorem}
$\soi$ as given by~\eqref{eq:soi} satisfies Condition \ref{cond:soi}.
\end{theorem}

In particular, the minimum of $\soi$ is 0 and achieved if and only if $\max(\binvars)=1$ for each one-hot constraint $\onehot(\binvars)$. 

Now, observe that:
\begin{align}
  \soi &= \sum_{o(\binvars) \in \O}\vio(\binvars) \nonumber= \sum_{o(\binvars) \in \O}(\min_{b\in\binvars}(1 - b)) \nonumber\\
   &= \min\Big(\big\{ f \mid f = \sum_{o(\binvars) \in \O} (1 - b_i), \quad b_i \in \binvars \big\}\Big).
  \label{eq:rearrange}
\end{align}
Thus, $\soi$ is the minimum over a set, which we denote $\soiset$, of linear functions. Although $\soi$ is concave piecewise-linear and cannot be directly minimized with convex optimization, we could minimize each individual function $f\in \soiset$ and take the minimum over all functions. Notice that each linear function in $\soiset$ has a semantic meaning: it corresponds to a particular mode sequence, i.e., a choice of modes at each time step. For notational convenience, we define $\costOf(f,\phi)$ to be the minimum of $f$ subject to $\phi$. Thus, the minimization problem \eqref{eq:min} can be restated as searching for a mode sequence $f\in \soiset$, where $\costOf(f,\rlx{\phi})$ is minimal.

\subsection{Minimizing the SoI with MCMC Sampling \label{subsec:minSOI}}
In the worst case, minimizing \soi requires enumerating and minimizing each mode sequence. In practice, the search can terminate once a mode sequence $f$ is found such that $\costOf(f,\rlx{\phi}) = 0$. A local search procedure that ``greedily'' navigates towards the minimum can be used to search for such mode sequence.  One such method is \emph{Markov chain Monte Carlo} (MCMC)~\cite{mcmc-1}, which can be viewed as a class of intelligent hill-climbing algorithms robust against local optima. In our context, MCMC can be used to generate a chain of mode sequences $f_0, f_1, f_2... \in \soiset$, with the desirable property that in the limit, the sampled mode sequences are more frequently from the minimum region of $\costOf(f,\rlx{\phi})$.

We use the Metropolis-Hastings (M-H) algorithm~\cite{mh}, a widely applicable MCMC method, to construct the sequence. The algorithm maintains a current mode sequence $f$ and proposes to replace $f$ with a new mode sequence $f'$. The proposal comes from a \emph{proposal distribution} $q(f' | f)$ and is accepted with a certain \emph{acceptance probability} $m(f{\rightarrow}f')$. If the proposal is accepted, $f'$ becomes the new current mode sequence. Otherwise, another proposal is considered. This process is repeated until one of the following scenarios happen:
1) a mode sequence $f$ where $\costOf(f,\rlx{\phi}) = 0$ is found;
2) a predetermined computational budget is exhausted; 
3) all possible mode sequences have been considered.

In order to employ the algorithm, we transform $\costOf(f,\rlx{\phi})$ into a probability
distribution $p(f)$ using a common method~\cite{mcmc}: $p(f) \propto \exp(-\beta \cdot \costOf(f,\rlx{\phi}))$,
where $\beta$ is a configurable parameter. We use the following acceptance probability
(often referred to as the \emph{Metropolis ratio})~\cite{mcmc}:
$m(f{\rightarrow}f') = \min(1, \frac{p(f')}{p(f)})$.

Importantly, under this acceptance probability, \textit{a proposal reducing the value of the cost function is always accepted, while a proposal that does not may still be accepted}
(with a probability that is inversely correlated with the increase in the cost). This means that the algorithm always greedily moves to a lower-cost mode sequence whenever it can, but it also has an effective means for escaping local minima.


\subsection{Stochastic Optimization for One-Hot Constraints}

\begin{algorithm}[t]
\small
\begin{algorithmic}[1]
\State {\bfseries Input:} $\C$, $\O$, $\E$
\State {\bfseries Output:} \feas/\infeas, feasible solution $\alpha$, theory lemmas $\L$ 
\State {\bfseries Parameters:} Sampling budget $T$ 
\Function{\deepsoiT}{$\C,\O, \E$}
\State $r, \alpha_0, \L \mapsto \convProc(\C\cup\rlx{\O})$
\If{$r = \infeas \lor \alpha_0\models \C\cup \O$} {\bf return} $r, \alpha_0, \L$
\EndIf
\State $k,f \mapsto 0, \initialCost(\alpha_0, \O)$ \label{line:initsoi}
\State $\alpha, c \mapsto \convOptProc(f, \C\cup\rlx{\O})$
\State {$\L \mapsto \{f > 0\}$}
\While{$c > 0 \land \neg \func{exhausted}() \land k < T$} 
\State $f' \mapsto \propose(f, \alpha, \E \cup \neg\prop(\L))$ \label{line:proposesoi}
\State $\alpha', c' \mapsto \convOptProc(f', \C\cup\rlx{\O})$
\If {$c' > 0$} {$\L \mapsto \L \cup \{f' > 0\}$}
\EndIf
\If {$\accept(c,c')$} $f,c, \alpha \mapsto f', c', \alpha'$
\Else {$\ k \mapsto k + 1$}
\EndIf
\EndWhile
\If {$\func{exhausted}() \land c > 0$} 
\State {\bf return} $\infeas, \alpha, \L$
\Else
\State {\bf return} $\feas, \alpha, \L$
\EndIf
\EndFunction
\end{algorithmic}
\caption{The \deepsoi procedure \label{alg:stoc}}
\end{algorithm}

The stochastic optimization procedure \deepsoi is shown in Alg.~\ref{alg:stoc}. It takes as input a set of linear constraints $\C$ and a set of one-hot constraints $\O$ and stochastically searches for a feasible solution to $\phi:= \C\cup\O$. It is intended as a drop-in replacement of the \convProc method in Line~\ref{line:checkconv} of Alg.~\ref{alg:complete} in order to more efficiently find solutions that satisfy not only the convex relaxation but also the binary constraints. Concretely, we will replace Line~\ref{line:checkconv} with 
\[
r, \alpha, \L \mapsto \deepsoi(\C\cup\D, \O, \E)
\]

\deepsoi follows the standard two-phase convex optimization approach. 
Phase I (Lines 5-6) finds a feasible solution $\alpha_0$ to $\rlx{\phi}$, and phase II (Lines 7-16) attempts to optimize \soi using the M-H algorithm. Phase II uses a standard convex optimization procedure $\convOptProc$ which takes an objective function $f$ and a set of convex constraints as inputs and returns a pair $(\alpha, c)$, where $\alpha\models\phi$ and $c = \costOf(f, \phi)$ is the optimal value of $f$. Phase II chooses an initial mode sequence $f$  based on $\alpha_0$ (Line~\ref{line:initsoi}) and computes its optimal value $c$. The M-H algorithm repeatedly proposes a new mode sequence $f'$ (Line~\ref{line:proposesoi}), computes its optimal value $c'$, and decides whether to accept $f'$ as the current mode sequence $f$. Moreover, if a mode sequence $f$ is found to be infeasible (e.g., $\costOf(f, \rlx{\phi}) > 0$), we record this as a theory lemma.

The procedure returns \infeas if the convex relaxation is infeasible (Line 6) or if the mode sequences are exhausted and none of them is feasible (Line 16). Otherwise, the procedure returns \feas. Moreover, when a mode sequence with cost 0 is found, the returned solution $\alpha$ is a feasible solution to the precise constraints $\C\cup\O$. Finally, the theory lemmas accumulated through the process are also returned for use by the main search algorithm.
Importantly, under the hood, the same convex optimization procedure is used in both phases. Therefore, from the perspective of the convex optimizer, \deepsoi solves a sequence of convex optimization problems that differ only in the objective functions, and each problem can be solved incrementally by updating the objective function without the need for a restart.

The $\accept$ method decides whether a proposal is accepted based on the Metropolis ratio. Function $\initialCost$ proposes the initial mode sequence $f$ by rounding the relaxed binary variables in $\alpha_0$ to the nearest integer. The $\propose$ method is more intricate, as explained below.

\subsection{Propagation-based proposal strategy}
\label{sec:proposal}

The proposal strategy for the M-H algorithm is key to its convergence efficiency. One popular proposal strategy was introduced in Walksat~\cite{selman1994noise}. In our context, it amounts to randomly changing the mode of a currently unsatisfied one-hot constraint. While this works reasonably well, there are two drawbacks: 1) it can get stuck in local optima (as one could cycle between mode sequences); 2) it does not take known theory lemmas into consideration. To tackle these two issues at once, we further leverage the fast propagation technology in the SAT solver to make sure that the proposed mode sequence is consistent with known propositional constraints. Our proposal strategy is described in Alg.~\ref{alg:proposal}.

\begin{algorithm}[t!]
\small
\begin{algorithmic}[1]
\State {\bfseries Input:} current mode sequence $f:=  \sum_{o(\binvars) \in \O} (1 - b_i), \quad b_i \in \binvars$, current solution $\alpha$, all propositional constraints $\E$
\State {\bfseries Output:} proposed mode sequence $f'$
\Function{propose}{$f, \alpha, \E$}
\State $\O' \mapsto \func{unsatisfied}(\O, \alpha)$
\State $o(\binvars^t) \mapsto \func{randomChoice}(\O')$
\State $b^t \mapsto \func{randomChoice}(\binvars^t \backslash \{b^t\})$
\State $Q \mapsto [\prop(b^t), \prop(b^1), \ldots \prop(b^{t-1}), \prop(b^{t+1}), \ldots, \prop(b^T)]$
\While {$\neg\satProc(\E \cup Q)$}
\State $Q.\pop()$
\EndWhile
{$\textbf{return } \func{satSolutionToModeSequence}()$}
\EndFunction
\end{algorithmic}
\caption{Propagation-based proposal strategy.\label{alg:proposal}}
\end{algorithm}

The procedure \propose takes as input the current mode sequence, the current solution $\alpha$ and all propositional constraints $\E$ (including all the theory lemmas found so far). Similar to Walksat, it first randomly selects a currently unsatisfied one-hot constraint (Lines 4-5) and randomly changes its mode to $b^t$ (Line 6). Then, instead of directly returning this ``adjacent'' mode sequence, the procedure uses a SAT solver to propagate the effect of this mode switch. Concretely, we first check whether the proposed mode combination is consistent with $\E$ (Line 8) using a SAT solver. If an inconsistency is detected, we leave one of the one-hot constraints unassigned (Line 9) and check the SAT-level consistency of the remaining partial mode combination. This process is repeated until we find a partial mode combination that is consistent with $\E$. At this point, we construct a full mode sequence setting the unassigned one-hot constraints according to the assignment found by the SAT solver.

\begin{theorem}
If $\E$ is satisfiable, Alg.~\ref{alg:proposal} always terminates with a proposed mode sequence that is consistent with $\E$.
\end{theorem}

In practice, the repeated invocation of $\satProc$ does not incur significant runtime overhead ($<5\%$ of the total runtime of \deepsoi). This overhead pays off in practice because it efficiently rules out infeasible mode sequences that would otherwise be attempted by convex optimization.
\section{\sys: An MILP Solver for PWA-Control}
\label{sec:overview}

We now present \sys, a specialized MILP solver for PWA control that combines the proposed techniques in Secs.~\ref{sec:dpllt} and \ref{sec:soi}. \sys is implemented in $\sim$15K lines of C++ code with 100+ unit tests. Fig.~\ref{fig:overview} exhibits its architectural design. 

\smallskip \noindent \textbf{Parser} \sys takes as input MILP problems in the standard \mps format, which are supported by most off-the-shelf MICP solvers and can be generated by popular robotic software such as Drake~\cite{drake}. This makes it relatively straightforward to use \sys or compare it with existing solvers. Our \mps parser transforms the given constraints into an internal representation (IR). A unique feature is that it will automatically extract \emph{one-hot constraints} from the \mps file and create for them both the standard arithmetic representation and a propositional logic level representation.

\smallskip \noindent \textbf{Pre-solver} Before entering the main solving loop described in Alg.~\ref{alg:complete}, \sys first performs interval analysis on both the linear constraints and the logical constraints. For example, if in a certain equation, all variables but one are bounded, the analysis will derive sound bounds for the unbounded variable. In practice, we find pre-solving can significantly reduce the runtime of the convex procedure.

\smallskip \noindent \textbf{Engine} After pre-solving, the updated IR is passed on to the main solving engine, which has three components, a SAT Solver, a convex solver, and the \deepsoi engine. The former two combined execute the DPLL(T) procedure described in Alg.~\ref{alg:complete}. We instantiate \satProc with the Cadical SAT solver~\cite{cadical}, and \convProc/\convOptProc with the LP solver in the Gurobi Optimizer~\cite{gurobi}, which is capable of generating explanations~\cite{gurobiexplanation}. Incremental solving is leveraged whenever possible.

There are two reasons we implement our own DPLL(T) solver rather than building on top of existing SMT solvers. First, LP engines in SMT-solvers use arbitrary-precision rational arithmetic, which can be many times slower than off-the-shelf floating-point LP solvers. While precise arithmetic is necessary for ensuring soundness in formal verification, it is not as crucial in our setting, where the goal is to \emph{find solutions}. Secondly, we hope to support more general MICP in the future, but support for convex programming beyond LP is less mature in SMT solvers. 

\section{Evaluation on Control of PWA Systems}
\label{sec:experiments}

We evaluate \sys on MILP encodings of PWA control problems. We compare \sys with two state-of-the-art MICP solvers -- \gurobi~\cite{gurobi} and \mosek~\cite{mosek} -- and perform ablation studies to show the effectiveness of our techniques. The main evaluation criterion is the run-time performance of finding a feasible solution (a trajectory).

\subsection{Solver configurations} 

We denote our best configuration as \best and compare with \gurobi and \mosek in their default configurations. Anecdotally, Gurobi has a particular strength in MILP, which is in fact the setting we consider. \emph{We remark that the relative performance of these solvers on the specific benchmarks we study may not be representative of their relative performance on other kinds of problems.} 

We run three ablation configurations, each of which differs from \best by one feature: 
\begin{enumerate*} [(1) ]
    \item \nocdcl does not extract theory lemmas from the convex procedure or perform logical reasoning with the SAT solver (i.e., Line \ref{line:checksat} of Alg.~\ref{sec:dpllt} is not executed). This configuration is a bare-bones branch-and-bound-like complete search that runs \deepsoi at each search state;
    \item \nosoi does not perform the \deepsoi procedure but runs a normal convex procedure during the search, making it less likely to find feasible solutions until it reaches the bottom of the search tree;
    \item \noprop uses the Walksat-based strategy in Sec.~\ref{sec:soi} rather than the propagation-based proposal strategy in \deepsoi.
\end{enumerate*}

\subsection{Benchmarks}

We evaluate on two types of existing PWA control problems, namely \emph{Stepping Stones}~\cite{marcucci2021shortest} and \emph{Ball and Paddle}~\cite{marcucci2019mixed}. 

\paragraph{Stepping Stones~\cite{marcucci2021shortest}}

\newcommand{\position}{\textbf{\textit{q}}}
\newcommand{\velocity}{\textbf{\textit{v}}}
\newcommand{\force}{\textbf{\textit{a}}}
\newcommand{\control}{\eta}
\newcommand{\timestamp}{t}
\newcommand{\timeMax}{T}

\begin{figure}
    \centering
        \begin{subfigure}[b]{0.70\linewidth}
            \includegraphics[height=2.5cm]{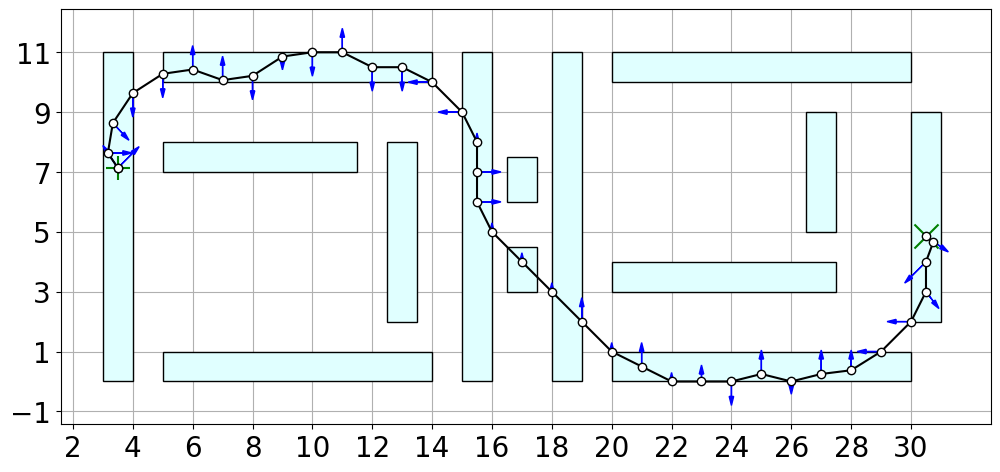}
            \caption{\ssa: 1st map}
        \end{subfigure}
        \hspace{-0.5cm}
        \begin{subfigure}[b]{0.28\linewidth}
            \includegraphics[height=2.55cm]{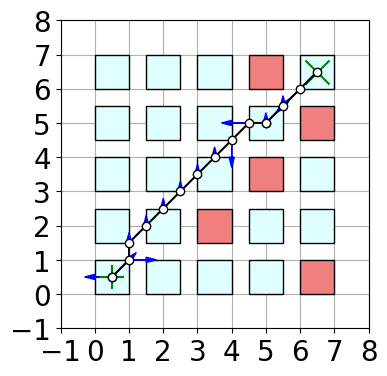}
            \caption{\ssb: 2nd map \label{fig:stepping_stone2}}
        \end{subfigure}
        \vspace{-2mm}
    \caption{Two types of maps with example solutions (planning trajectories from \sys) in the Stepping Stones benchmark.}
    \label{fig:stepping_stone}\vspace{-5mm}
\end{figure}

In these problems, an agent must navigate from a starting position to a destination, stepping only on the (convex) blue and red regions, as illustrated in Fig.~\ref{fig:stepping_stone}. We use the same dynamics as~\cite{marcucci2021shortest}, but design two new maps, \ssa and \ssb. The goal is to reach the destination position with zero velocity within a fixed number of time steps. The white regions are not reachable, while the blue and red regions have different costs. The one-hot constraint in this case is used to specify the fact that the agent is in one of the convex regions, and the number of modes is equal to the number of convex regions.
This problem can be seen as an abstraction of the footstep planning scenario in Fig.~\ref{fig:motivation}, although we remark that the latter requires quadratic constraints, so the problem becomes MIQP rather than MILP. Supporting MIQP in \sys is a direction for future work.

We generated $100$ MILP encoding instances from map \ssa by randomly varying the vertical positions of the starting and ending points in $[0, 11] \times [2, 9]$, respectively. We generated another $100$ MILP problems from map \ssb by randomly selecting $5$ (red regions in Fig~\ref{fig:stepping_stone2}) out of the $16$ convex regions to have a lower control parameter, which makes navigation harder. We refer to the benchmark sets as \ssa and \ssb in Table~\ref{tab:main}.

\paragraph{Ball and Paddle~\cite{marcucci2019mixed}}

\newcommand{\paddle}{\textbf{\textit{u}}}

\begin{figure}
    \centering
    \includegraphics[width=0.22\linewidth]{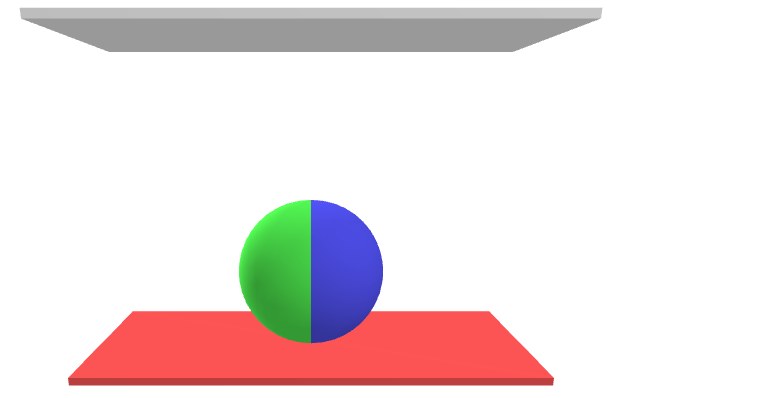}
    \includegraphics[width=0.22\linewidth]{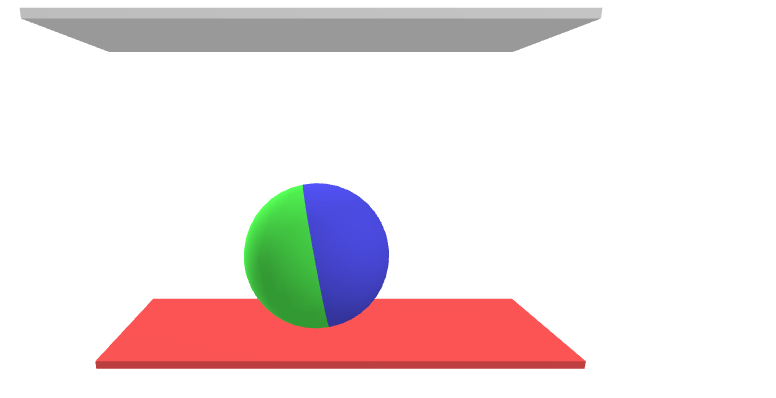}
    \includegraphics[width=0.22\linewidth]{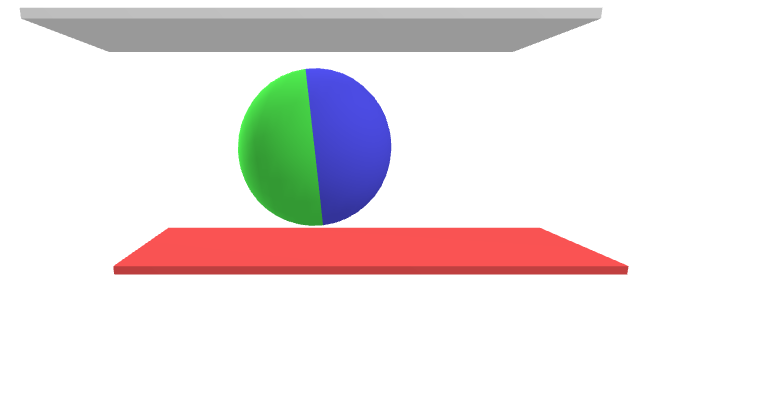}
    \includegraphics[width=0.22\linewidth]{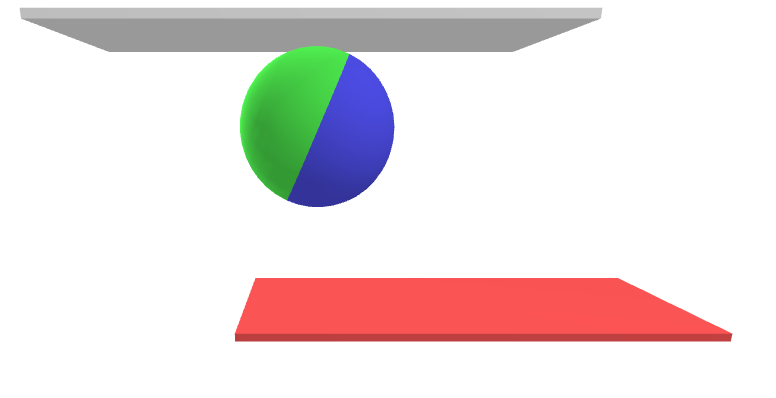}
    \includegraphics[width=0.22\linewidth]{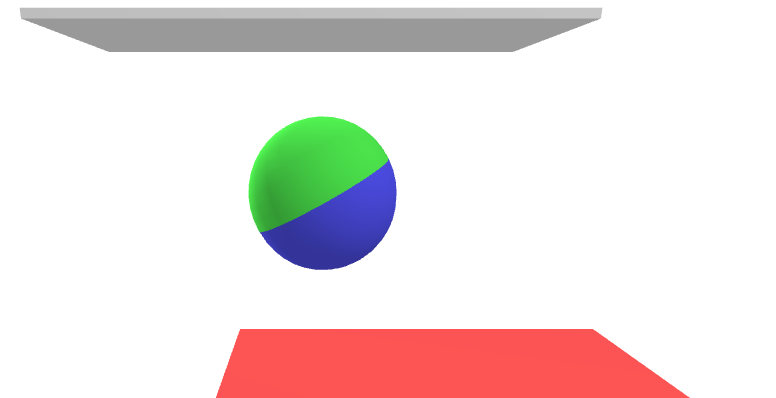}
    \includegraphics[width=0.22\linewidth]{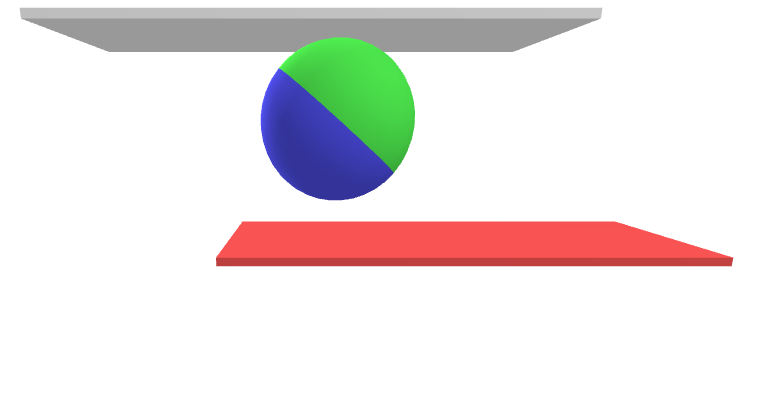}
    \includegraphics[width=0.22\linewidth]{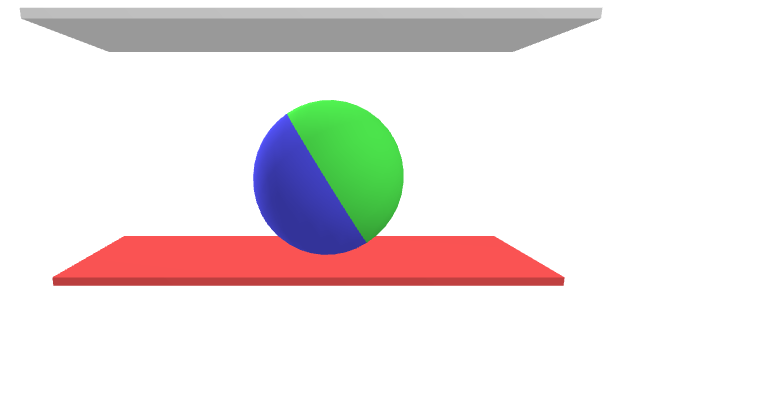}
    \includegraphics[width=0.22\linewidth]{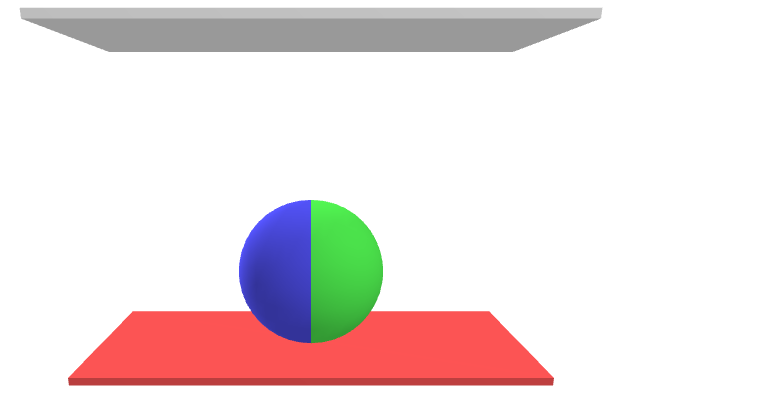}
    \caption{Flipping the ball by $180\degree$ with a paddle. The top right image is the \emph{initial} state and the bottom right is the \emph{terminal} state. These images are sampled frames from the animation video in the supplementary materials.}
    \label{fig:ball_paddle}
    \vspace{-0.7cm}
\end{figure}

This problem, as shown in Fig.~\ref{fig:ball_paddle}, is to rotate a two-dimensional ball using a paddle (bottom in red) under a fixed ceiling (top in gray). 
The original system has 7 modes: no contact, ball in contact with the paddle (sticking, sliding left or right), and ball in contact with the ceiling (sticking, sliding left or right). We fix the terminal state and generate different instances by varying the initial state. The first benchmark set, denoted as \bpeasy, is generated by randomly varying the horizontal position of the ball in the interval $[0, l/2]$, where $l$ is the width of the paddle. The second benchmark set, denoted as \bphard, is from randomly varying the horizontal position and the angle of the ball in the interval $[0, l]\times [-\pi, \pi]$. Each benchmark set contains $100$ instances.

\paragraph{Experimental setup}
Each benchmark is stored in MPS format. Each configuration was given one thread and a 20 minute CPU-timeout per benchmark. Experiments were performed on a cluster equipped with Intel Xeon E5-2637 v4 CPUs running Ubuntu 20.04. We measure the CPU time each configuration takes on each instance. The instances that timed out are assigned the time limit as their running time. 

\subsection{Comparison with existing solvers}

\begin{table*}[t!]
\setlength\tabcolsep{5pt}
\begin{center}
\renewcommand{\arraystretch}{1.1}
\begin{tabular}{cccccccccccc}
& & \multicolumn{2}{c}{\ssa} & \multicolumn{2}{c}{\ssb} & \multicolumn{2}{c}{\bpeasy} & \multicolumn{2}{c}{\bphard} & \multicolumn{2}{c}{Total} \\ 
\toprule
& \textbf{Configuration}  & Time   & \# Solved & Time & \# Solved & Time & \# Solved & Time & \# Solved & Time   & \# Solved \\
\cmidrule(lr){2-2} \cmidrule(lr){3-4} \cmidrule(lr){5-6} \cmidrule(lr){7-8}  \cmidrule(lr){9-10} \cmidrule(lr){11-12} 
\multirow{3}{*}{\bf{Main}} & \gurobi             & \textbf{868}    &    100   &  \textbf{212}  &  100    & 3465 &      100  &  8803 &     99 & 13348  & 399 \\
&\mosek              & 41683  &    75    & 1666 &   100    & 109373 &      13 & 111053 &    14  & 263775 & 202 \\
&\textbf{\best}              & 1057    &   100    & 458 &   100    & \textbf{1488} &  100 & \textbf{2757}  &  100 & \textbf{6367} & \textbf{400}  \\ \hline
\multirow{2}{*}{\bf{Virtual}} & \gurobiandbest & 738 & 100 & 201 & 100 & 1205 & 100 & 2387 & 100 & 4531 & 400 \\
&   \virtualall  & 729 & 100 & 200 & 100 & 1205 & 100 & 2387 & 100 & 4521 & 400 \\ 
\hline
\multirow{3}{*}{\bf{Ablation}} & \nocdcl & 1430 & 100 & 533 & 100 & 37165 & 74 & 58282 & 55 & 97410 & 329  \\
&  \nosoi  & 26709 & 89 & 549 & 100 & 118957 & 5 & 109658 & 9 & 255873 & 203 \\
&  \noprop & 1207 & 100 & 636 & 100 & 2670 & 100 & 4642 & 100 & 9155 & 400 \\
\bottomrule
\end{tabular}
\end{center}
\caption{Runtime performance (in seconds) of different solver configurations on the benchmarks. The configuration are divided into the \textbf{main} configurations, the \textbf{virtual} portfolio configurations, and the \textbf{ablation} configurations.} \label{tab:main}
\vspace{-4mm}
\end{table*}

The runtime performance of all configurations running on all benchmarks is shown in Table~\ref{tab:main}. 
We first compare \best with \gurobi and \mosek (the \textbf{Main} block). On the Stepping Stones benchmarks (the \ssa and \ssb columns), both \best and \gurobi can efficiently solve all the instances, while \mosek fails to solve 25 instances within the time limitation. \gurobi is overall faster on those benchmarks, with an average runtime of 8.7 seconds per instance on \ssa and 2.12 seconds per instance on \ssb; in contrast, \best has an average runtime of 10.6 and 4.6 seconds, respectively. 

The Ball and Paddle benchmarks take significantly more time to solve. \mosek struggles in this more challenging use case and only solves 27 instances. On the other hand, \best performs significantly better than \gurobi. Not only does it take 57\% less time on \bpeasy and 69\% less time on \bphard, it also solves all of the 200 instances while Gurobi times out on 1 of them. Running with a longer time limit reveals that \textbf{it takes \gurobi 7 hours and 39 minutes to find a feasible solution for this instance}. In contrast, \best solves this instance in 501 seconds. 

\begin{figure}[h!]
    \centering
    \includegraphics[width=0.66\linewidth]{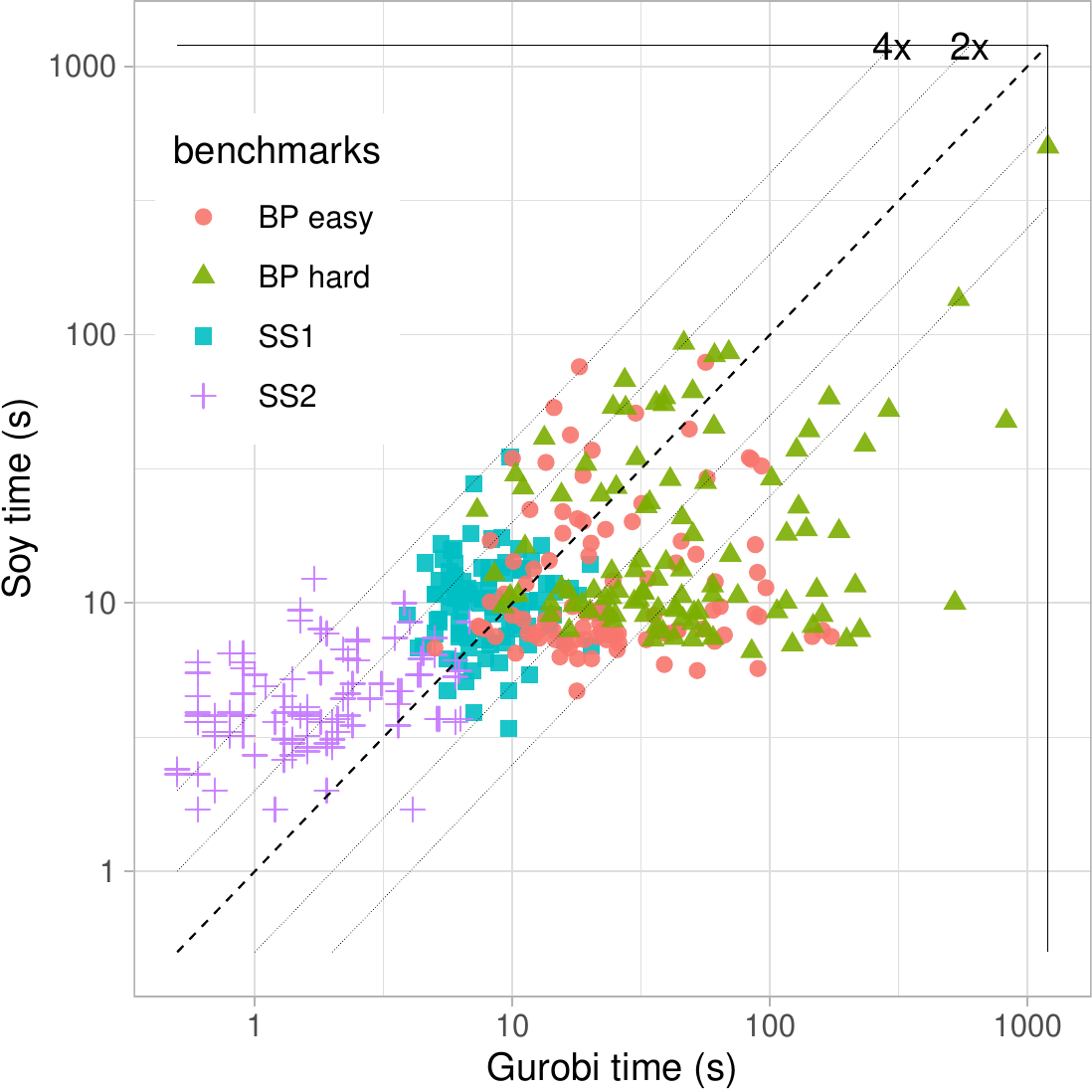}
    \caption{Scatter plot of the runtime of \best and \gurobi.}
    \vspace{-2mm}
    \label{fig:scatter}
    \vspace{-4mm}
\end{figure}

A scatter plot of the runtime of \best and \gurobi on all benchmarks is shown in Fig.~\ref{fig:scatter}. Overall, \best performs best, but the two solvers frequently show complementary behaviors. Indeed, as shown in Table~\ref{tab:main}, if we consider a virtual portfolio strategy that runs \best and \gurobi in parallel (\gurobiandbest), further performance gain can be obtained on \emph{all} of the four benchmark sets. This suggests that running multiple competitive solvers in parallel is advisable in practice. Marginal gain can be obtained if we also include \mosek in the portfolio (\virtualall).

\subsection{Ablation studies}

The \textbf{Ablation} block of Tab.~\ref{tab:main} shows the runtime performance of the ablation configurations. The number of solved instances drops significantly if we turn off logical reasoning (\nocdcl) or if we do not perform \deepsoi (\nosoi). Without \deepsoi, the performance of \best is disastrous on the Ball and Paddle benchmarks, solving only 14 of the 200 instances. Interestingly, we observe a similar pattern on the performance of \mosek and \nosoi: they both fail on a fraction of \ssa, solve all of \ssb, and perform badly on \bpeasy and \bphard. This leads us to speculate that \mosek does not invest in local search during MILP solving. Finally, if we use a Walksat-like proposal strategy instead of the propagation-based proposal strategy in the \deepsoi procedure, we are still able to solve all instances, but the overall runtime degrades by 44\% (from 6367 to 9155 seconds).  The conclusion of these ablation studies is that each of the proposed techniques contributes to the competitive performance of \best.

\section{Conclusion and Future Directions}
\label{sec:next-steps}

We have introduced \sys, a specialized MILP solver for PWA control problems. We instantiated the DPLL(T) procedure for deciding the feasibility of combinations of linear and one-hot constraints. We also presented \deepsoi, a specialized optimization procedure that stochastically minimizes the sum-of-infeasibilities to search for feasible mode sequences. \sys is already competitive against highly-optimized off-the-shelf MICP solvers, suggesting that this direction of designing specialized MICP solvers for PWA control is promising. We hope the work will garner interest in \sys and welcome contributions of new benchmarks for \sys. 

\smallskip\noindent \textbf{Limitations and Future Work.} \sys is still a research prototype, and as such has many limitations compared to mature solvers: \begin{enumerate*} [1) ]
    \item We do not yet support more general convex constraints, such as quadratic constraints, which many MICP applications (e.g., \cite{deits2014footstep}) require; 
    \item We do not yet support other logical constraints common in robotic applications (e.g., ``at-least-one'', ``at-most-one'');
    \item We only perform feasibility checks instead of finding optimal solutions.
\end{enumerate*}

We hope to address all these limitations in the future. We note that the proposed techniques are compatible with more general convex constraints and can be in principle extended to other logical constraints, though actually doing these efficiently requires non-trivial research and engineering effort. On the other hand, a strong feasibility checker is the first step towards a strong optimizer. In the SMT community, there is an active effort to extend SMT solvers to do optimization (a line of work called ``optimization modulo theories'')~\cite{sebastiani2020optimathsat} and insights there could potentially be borrowed.

Supporting parallelism, incrementality, and APIs to encode constraints are also important next steps for our tool.

\smallskip\noindent \textbf{Acknowledgment.} This work was partially supported by grants from the National Science Foundation (2211505 and 2218760) and by the Stanford Center for AI Safety.
We thank Tobia Marcucci,
Gustavo Araya,
Richard McDaniel,
J\"org Hoffmann,
Aina Niemetz,
Mathias Preiner,
and Makai Mann for the helpful discussions.

\bibliographystyle{IEEEtranS}

\bibliography{bibli}

\end{document}